\documentclass{article} 
\usepackage{iclr2027_conference,times}

\usepackage{amsmath,amsfonts,bm}

\def\eqref#1{equation~\ref{#1}}

\def\1{\bm{1}}

\DeclareMathAlphabet{\mathsfit}{\encodingdefault}{\sfdefault}{m}{sl}
\SetMathAlphabet{\mathsfit}{bold}{\encodingdefault}{\sfdefault}{bx}{n}

\usepackage{hyperref}
\usepackage{url}

\usepackage{amsmath}
\usepackage{amssymb}
\usepackage{mathtools}
\usepackage{amsthm}

\usepackage[table]{xcolor}
\usepackage{multirow}
\usepackage{makecell}
\usepackage{pifont}       
\usepackage{bbding}      
\usepackage{fontawesome}
\usepackage{subcaption}
\usepackage{arydshln}
\usepackage{tabularx}
\usepackage{mdframed}
\usepackage[most]{tcolorbox}
\usepackage{arydshln}
\usepackage{adjustbox}
\usepackage{wrapfig}
\usepackage{titletoc}
\usepackage{xspace}
\usepackage{booktabs,array,colortbl,graphicx}

\definecolor{table-blue}{RGB}{173, 216, 230}
\definecolor{row-highlight}{RGB}{220, 230, 242}
\definecolor{qwen-header}{RGB}{235, 240, 248}
\definecolor{llama-header}{RGB}{248, 244, 235}
\definecolor{section-gray}{RGB}{245, 245, 245}
\definecolor{group-header}{RGB}{230, 235, 245}

\definecolor{zhz_gray}{rgb}{0.8,0.8,0.8}
\definecolor{darkgreen}{rgb}{0.0, 0.5, 0.0} 
\definecolor{darkred}{rgb}{0.5, 0.0, 0.0}   

\definecolor{row-alt}{RGB}{248, 250, 253}      
\definecolor{hl-llama}{RGB}{234, 242, 255}     
\definecolor{hl-qwen}{RGB}{236, 250, 240}      

\definecolor{datasetbg}{HTML}{E8EDF5}
\definecolor{perfbg}{HTML}{EAF3FF}
\definecolor{costbg}{HTML}{FFF1E3}
\definecolor{oursbg}{HTML}{EEF8F0}
\definecolor{groupbg}{HTML}{F5F5F5}

\input{components}

\newcommand{\hivemem}{\textsc{HiVe-Mem}\xspace}

\title{\hivemem: Interleaving \textbf{\textsc{H}}or\textbf{\textsc{i}}zontal Memory Expansion and \textbf{\textsc{Ve}}rtical Evidence Composition for Multimodal Agents}

\author{
  Haozhen Zhang$^1$\thanks{Equal Contribution. \quad Corresponding: \texttt{wazhz14@gmail.com}}, 
  ~Bohan Liu$^{1*}$, 
  ~Jianzhu Bao$^1$,
  ~Qingyuan Liu$^1$,
  ~Haodong Yue$^2$,
  ~Xiao Li$^3$, \\
  {\bf
  ~Zi Pong Lim$^4$,
  ~Wenya Wang$^1$
  } \\
  $^1$Nanyang Technological University ~
  $^2$Tsinghua University ~
  $^3$Sun Yat-sen University ~\\
  $^4$Aumovio Singapore Pte. Ltd.\\
}

\iclrfinalcopy 
\begin{document}

\maketitle

\begin{abstract}

As multimodal agents accumulate increasingly long interaction histories, effectively retrieving and utilizing past information becomes critical for long-term interaction.
Existing memory systems typically rely on semantic retrieval over compressed memories and expose predefined representations, which may miss relevant memories and overlook heterogeneous evidence needs across queries.
We introduce \textbf{\hivemem}, a multimodal memory framework that interleaves \textbf{\textsc{H}}or\textbf{\textsc{i}}zontal Memory Expansion with \textbf{\textsc{Ve}}rtical Evidence Composition.
For horizontal memory expansion, \hivemem organizes memory episodes into a relational graph using textual and visual anchors, expanding semantically retrieved seeds with related neighbors to recover complementary information.
For vertical evidence composition, each retrieved memory retains multiple evidence forms, including compact summaries, captions, grounded visual views, raw images, and interaction history.
A lightweight trainable selector independently composes an evidence subset for each retrieved memory, enabling query-adaptive utilization of heterogeneous multimodal information.
This two-dimensional design jointly broadens memory coverage and adapts the evidence contributed by individual memories.
Experiments across three multimodal memory benchmarks show that \hivemem consistently improves task performance across base models while reducing token consumption by about 90\% compared with the most efficient competing multimodal baselines.
Code is available at \href{https://github.com/ViktorAxelsen/HiVe-Mem}{https://github.com/ViktorAxelsen/HiVe-Mem}

\end{abstract}

\section{Introduction}

As multimodal agents increasingly engage in persistent interactions involving both language and visual observations, memory becomes essential for preserving and reusing information from past interactions over long horizons~\citep{hu2025memory, huang2026rethinking}. 
Existing memory systems typically transform interaction histories into compact representations and retrieve relevant entries when needed, allowing agents to access past information beyond the immediate context~\citep{kang2025memory, xu2025mem, fang2025lightmem, sun2023simple}. 
In multimodal settings, however, useful information can span multiple modalities and levels of granularity~\citep{bei2026mem, liu2026worldmemarena}, making it difficult for a fixed memory representation and retrieval process to consistently surface the information most useful for future queries~\citep{liu2026omni, wang2025mirix, long2026seeing}. 
This raises a fundamental challenge: \emph{how can multimodal memory systems flexibly identify and utilize relevant information from heterogeneous interaction histories at inference time?}

Despite recent progress, existing multimodal memory systems still face two key limitations.
\emph{First, structured memory is often centered on abstract entities and relations rather than interaction episodes themselves}.
While such structures are effective for consolidating knowledge, they may fail to capture cross-episode contextual affinities in multimodal histories, where related episodes can share partial cues across text and images, such as entities, scenes, or events, without forming explicit symbolic relations~\citep{liu2025memverse, liu2026omni}.
As a result, retrieval may still miss relevant episodes beyond directly matched memories, especially when useful information is distributed across multiple interactions.
\emph{Second, retrieved memories are typically exposed through predefined representations}, despite their heterogeneous evidence requirements.
While compact textual abstractions may suffice for some memories, others require fine-grained visual or contextual evidence, making a fixed multimodal representation either insufficient or unnecessarily detailed, thereby limiting memory utilization and increasing token consumption~\citep{long2026seeing, wang2025mirix}.
These limitations call for a memory framework that can both recover relevant information distributed throughout multimodal histories and adaptively compose the evidence exposed by each retrieved memory.

To address these challenges, we propose \textbf{\hivemem}, which broadens semantic retrieval through \emph{Horizontal Memory Expansion} and tailors the multimodal representation of each retrieved memory through \emph{Vertical Evidence Composition}.
Horizontally, \hivemem organizes memory episodes into a lightweight relational graph using textual and visual anchors extracted from multimodal histories.
Starting from semantically retrieved seeds, it expands retrieval toward associated episodes through cross-episode contextual affinities, complementing direct semantic matching with relational signals.
Vertically, \hivemem retains multiple evidence forms for each memory, including compact summaries, captions, grounded visual views, raw images, and interaction history.
A lightweight RL-trained selector independently composes the evidence for each retrieved memory conditioned on the query, allowing different memories to expose different evidence subsets while jointly considering answer quality and token consumption.
Together, these two dimensions broaden relevant recall while adapting multimodal evidence to the needs of individual memories.
Experiments across three multimodal memory benchmarks and two base models show consistent improvements over existing multimodal methods with substantially lower token consumption.
Further analyses validate the complementary benefits of horizontal expansion and adaptive evidence composition, demonstrate flexible behavior across query and budget settings, and show only marginal runtime overhead.

Our main contributions are summarized as follows:
\begin{itemize}
\item We propose \textbf{\hivemem}, a multimodal memory framework that jointly improves memory retrieval and evidence utilization, enabling flexible use of long text--image interaction histories beyond fixed retrieval and representation pipelines.
\item We realize this framework via complementary mechanisms. \emph{Horizontal Memory Expansion} augments semantic retrieval with cross-episode contextual affinities, while \emph{Vertical Evidence Composition} adaptively composes multimodal evidence per retrieved memory.
\item Extensive experiments across three multimodal memory benchmarks and two base models demonstrate improvements over multimodal methods with substantially lower token consumption. Further analyses validate both components and flexible evidence adaptation.
\end{itemize}

\section{Related Work}

\subsection{Text-Centric Agent Memory}

LLM agents increasingly rely on external memory to preserve and reuse information beyond bounded context windows~\citep{zhong2024memorybank,packer2023memgpt}. A major line of work transforms interaction histories into compact and reusable representations through extraction, compression, and consolidation~\citep{chhikara2025mem0,xu2025mem,kang2025memory,fang2025lightmem,liu2026simplemem}. Beyond independent memory entries, structured approaches organize past experiences through entities, relations, or associative links, allowing retrieval to exploit dependencies among memories rather than relying solely on isolated similarity matching~\citep{zhu2026hela}. Another line introduces learned memory management, where policies or evolving operations adapt how memories are constructed, updated, and reused from downstream feedback~\citep{yan2025memory,zhang2026memskill,yu2026agentic,wang2025mem,zhang2026learning}. Together, these methods have substantially advanced long-term memory for text-based agents, but remain primarily designed around textual information, leaving heterogeneous multimodal evidence less explored.

\subsection{Multimodal Agent Memory}

Recent work extends agent memory beyond text to persistent visual and multimodal interaction histories. MIRIX~\citep{wang2025mirix} maintains specialized memory types for heterogeneous user information, while M3-Agent~\citep{long2026seeing} organizes episodic and semantic experience around entity-centric multimodal representations. M2A~\citep{feng2026m2a} complements raw interaction records with higher-level semantic memories, whereas MemVerse~\citep{liu2025memverse} consolidates multimodal experience into hierarchical knowledge structures. Recent systems further explore contextual memory organization, reliability-aware retrieval, and unified multimodal memory pipelines~\citep{jain2025augustus,lu2026mma,liu2026omni}. Collectively, these methods establish effective mechanisms for storing, structuring, and accessing multimodal experience. In contrast, \hivemem focuses on improving how constructed memories are utilized at inference time, by broadening retrieval through relations among memory episodes and adaptively composing the evidence exposed by each retrieved memory, enabling more effective multimodal memory use.

\section{Methodology}

\begin{figure}[t]
    \centering
    \includegraphics[width=\linewidth]{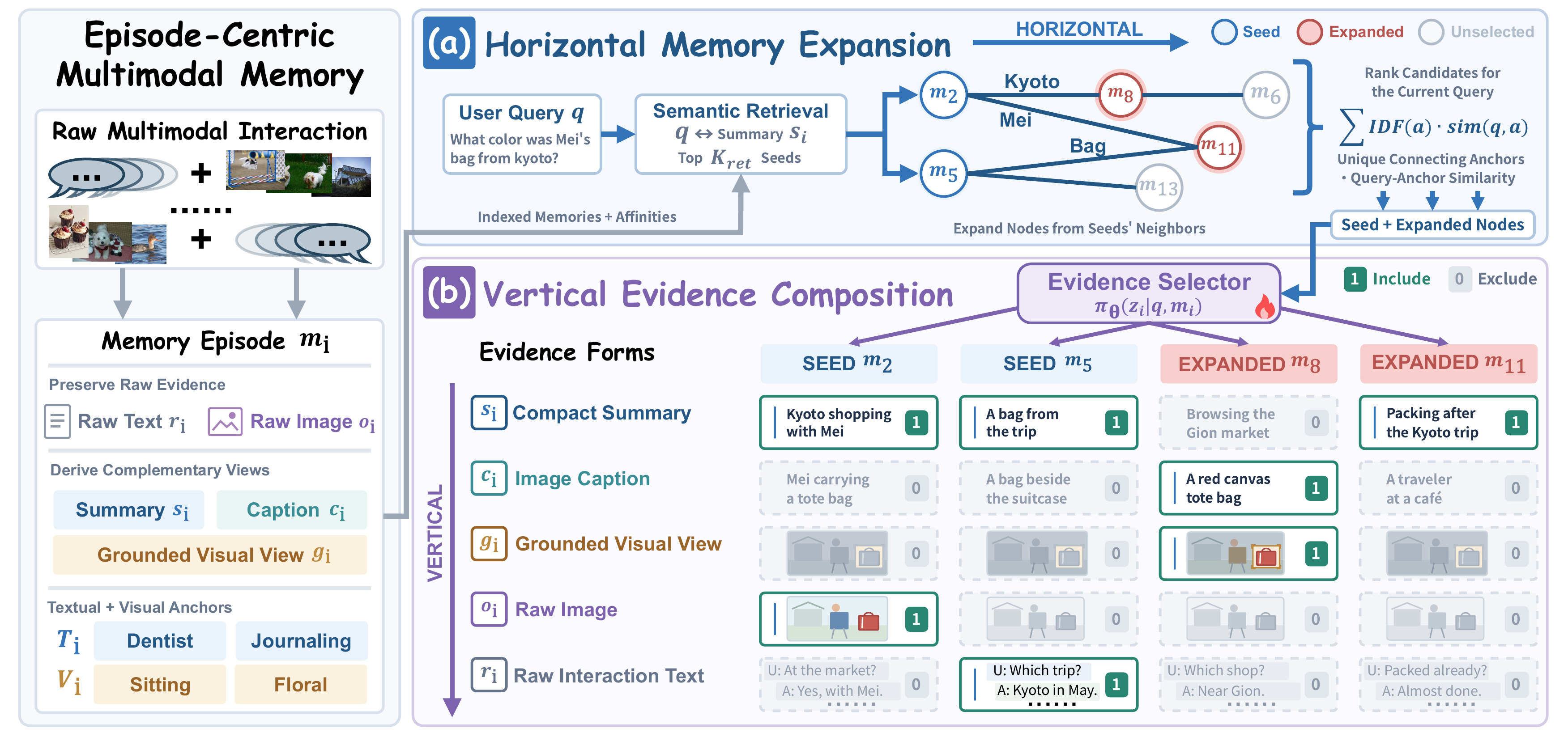}
    \caption{
Overview of \hivemem.
\textbf{(a) Horizontal Memory Expansion} retrieves seed episodes through summary-based semantic matching and expands to graph neighbors ranked by the specificity and query relevance of their connecting anchors.
\textbf{(b) Vertical Evidence Composition} uses an RL-trained selector to make binary inclusion decisions over compact summaries, image captions, grounded visual views, raw images, and raw interaction text, composing a query-dependent evidence subset for each retrieved episode to balance answer quality and token consumption.
}
    \label{fig:hivemem}
\end{figure}

Our methodology consists of three stages as shown in Figure~\ref{fig:hivemem}.
We first organize the interaction history into episode-centric memory representations with complementary textual and visual information in Sec.~\ref{sec:memory_formation}.
We then introduce \emph{Horizontal Memory Expansion} in Sec.~\ref{sec:horizontal} to complement semantic retrieval with cross-episode contextual associations.
Finally, Sec.~\ref{sec:vertical} presents \emph{Vertical Evidence Composition}, which adaptively determines the evidence exposed by each retrieved memory.

\subsection{Episode-Centric Multimodal Memory Formation}
\label{sec:memory_formation}

We consider a growing interaction history
$\mathcal{H}=\{x_t\}_{t=1}^{T}$, where each interaction may contain textual content and associated images.
Rather than treating every turn independently, we partition the history into contiguous \emph{interaction episodes} (formed over fixed-length token spans), offering a practical granularity while preserving local context.
Each episode $e_i$ is subsequently transformed into a \emph{memory episode} $m_i$, which serves as the basic memory object in \hivemem.

To instantiate each memory episode, we derive a compact summary $s_i$ with vision-language models (VLMs), providing a concise semantic representation for subsequent retrieval.
To preserve salient information beyond this coarse abstraction, we associate each episode with two sets of fine-grained, VLM-generated modality-aware anchors that capture complementary cues from language and vision.
Specifically, the \emph{textual anchors} $\mathcal{T}_i$ capture salient semantics from the interaction text, such as entities, events, and topics, while the \emph{visual anchors} $\mathcal{V}_i$ record visually grounded cues, such as objects, scenes, activities, and attributes, from the associated images.
Visual anchors are useful for retaining visually salient information that may be only partially captured by textual abstractions.
Together, the two anchor sets provide lightweight modality-specific metadata beyond the summary and are used to characterize contextual affinities between memory episodes in Sec.~\ref{sec:horizontal}.

Beyond these compact descriptors, each memory episode retains multiple evidence forms at different levels of detail.
By default, these include the raw interaction text $r_i$, compact summary $s_i$, image captions $c_i$, raw images $o_i$, and a grounded visual view $g_i$, which we introduce to localize context-relevant regions in the original images while suppressing irrelevant and noisy background, inspired by~\citep{liu2026seeing}.
Accordingly, each memory episode is represented as
\begin{equation}
m_i =
\big(
r_i,\,
s_i,\,
c_i,\,
o_i,\,
g_i,\,
\mathcal{T}_i,\,
\mathcal{V}_i
\big).
\end{equation}

This episode-centric representation retains complementary views of multimodal interaction history rather than reducing each memory to a single abstraction.
The compact summary supports semantic retrieval, the modality-aware anchors provide cues for modeling cross-episode affinities, and the evidence forms preserve complementary textual and visual details.
Together, these components support the \emph{Horizontal Memory Expansion} and \emph{Vertical Evidence Composition} introduced next.

\subsection{Horizontal Memory Expansion}
\label{sec:horizontal}

Semantic retrieval is effective at identifying memories directly aligned with the current query, but it largely considers each memory in isolation. In multimodal histories, relevant context may also be connected through weaker cues distributed across text and images, while explicit entity--relation structures primarily capture symbolic relations. To complement these signals, we model \emph{cross-episode contextual affinities} between memory episodes. A graph provides a natural substrate for preserving such soft associations without requiring predefined relation types, allowing retrieved memories to serve as entry points for discovering additional relevant context.

\textbf{Cross-Episode Affinity Graph.}
We construct an undirected affinity graph $\mathcal{G}=(\mathcal{M},\mathcal{E})$, where each node corresponds to a memory episode $m_i$.
We characterize pairwise affinity through overlaps among the textual and visual anchors defined in Sec.~\ref{sec:memory_formation}.
These anchors preserve fine-grained semantic cues beyond summary-level similarity, particularly cross-modal correspondences where a concept is expressed textually in one episode but appears visually in another.
For two memory episodes $m_i$ and $m_j$, we define their matching anchor set as
\begin{equation}
\mathcal{A}_{ij}
=
(\mathcal{T}_i \cap \mathcal{T}_j)
\cup
(\mathcal{T}_i \cap \mathcal{V}_j)
\cup
(\mathcal{V}_i \cap \mathcal{T}_j).
\end{equation}
The first term captures shared textual semantics, while the latter two form cross-modal bridges between language and vision.
To reflect the varying specificity of matched anchors, we weight them by inverse document frequency (IDF), which suppresses frequent cues while emphasizing more distinctive ones. For each $a\in\mathcal{A}_{ij}$, with $N$ memory episodes, we define
\begin{equation}
\mathrm{df}(a)
=
\sum_{k=1}^{N}
\mathbb{I}\!\left[a\in\mathcal{T}_k\cup\mathcal{V}_k\right],
\qquad
\mathrm{IDF}(a)
=
\log\frac{N+1}{1+\mathrm{df}(a)},
\qquad
w_{ij}
=
\sum_{a\in\mathcal{A}_{ij}}
\mathrm{IDF}(a).
\end{equation}
The resulting $w_{ij}$ emphasizes more distinctive shared anchors while downweighting ubiquitous ones. To keep the graph sparse, we retain at most $K_{\mathrm{edge}}$ edges per node by greedily selecting candidate edges in descending $w_{ij}$ order while respecting the degree limit.

\textbf{Affinity-Guided Memory Expansion.}
Given a query $q$, we first retrieve $K_{\mathrm{ret}}$ seed memories by measuring semantic similarity between $q$ and each compact summary $s_i$, and collect their one-hop neighbors, excluding already retrieved memories, as expansion candidates.
To prioritize neighbors based on query-relevant anchor overlap, we score the fine-grained anchors they share with the seed memories rather than reuse summary-level similarity.
For a candidate $m_u$, let $\mathcal{B}_u$ denote the unique anchors shared with any seed memory.
Its expansion score is defined as
\begin{equation}
S_{\mathrm{exp}}(m_u \mid q)
=
\sum_{a\in\mathcal{B}_u}
\mathrm{IDF}(a)\,
\operatorname{sim}(q,a).
\end{equation}
where $\operatorname{sim}(q,a)$ denotes the cosine similarity between the query and anchor embeddings. The IDF term captures anchor specificity, while the similarity term measures query relevance. We rank expansion candidates by $S_{\mathrm{exp}}$ and select the top $K_{\mathrm{exp}}$, yielding up to $K_{\mathrm{ret}}+K_{\mathrm{exp}}$ retrieved memory episodes that combine direct semantic relevance with query-relevant cross-episode affinities.

\subsection{Vertical Evidence Composition}
\label{sec:vertical}

After horizontal expansion determines which memory episodes to retrieve, the remaining question is how each retrieved memory should be presented to the downstream model.
A single evidence form is unlikely to suit all memories in multimodal settings.
Compact summaries and captions are efficient but may omit fine-grained details, while raw interaction text and images preserve richer information at higher inference cost.
Moreover, the appropriate evidence can vary across both queries and retrieved memories.
We therefore formulate memory utilization as \emph{per-memory evidence composition}, where each retrieved memory contributes a query-dependent subset of its available evidence.

\textbf{Query-Adaptive Evidence Composition.}
For each retrieved memory $m_i$, we consider candidate evidence forms
$\mathcal{E}_i=\{r_i,s_i,c_i,o_i,g_i\}$ by default, corresponding to raw interaction text, compact summary, image caption, raw image, and grounded visual view (note that this formulation is not limited to these forms and can accommodate additional evidence types).
Rather than selecting a single representation, we introduce a lightweight trainable selector to jointly determine which evidence forms should be exposed.
For each memory, the selector takes the concatenated embeddings of the query and compact summary,
$\mathbf{h}_i=[E(q);E(s_i)]$, as its lightweight input.
A single forward pass jointly predicts evidence-specific selection probabilities
\begin{equation}
\mathbf{p}_i
=
f_{\theta}(\mathbf{h}_i)
=
(p_{i,1},\ldots,p_{i,|\mathcal{E}_i|})
\in [0,1]^{|\mathcal{E}_i|},
\end{equation}
where $p_{i,k}$ denotes the selection probability of the $k$-th evidence form.
These probabilities parameterize binary decisions, collected as
$\mathbf{z}_i=(z_{i,1},\ldots,z_{i,|\mathcal{E}_i|})$,
where $z_{i,k}\in\{0,1\}$ indicates whether the corresponding evidence form is included.
Applying the selector to each retrieved memory allows different episodes to expose different evidence subsets for the same query.

\textbf{Learning Evidence Composition.}
Determining the optimal evidence subset is non-trivial because its utility is ultimately reflected by downstream answer quality, while the selection itself is discrete.
We therefore optimize the selector with reinforcement learning~\citep{schulman2017proximal} using a reward that jointly considers answer quality and evidence consumption.
Specifically, for the composed retrieval context and resulting answer $\hat{y}$, we define
\begin{equation}
R
=
R_{\mathrm{ans}}(\hat{y},y)
-
\lambda \widetilde{C}(\mathbf{z}),
\qquad
C(\mathbf{z})
=
\sum_i\sum_k z_{i,k}\,\tau_{i,k},
\qquad
\widetilde{C}(\mathbf{z})
=
\operatorname{Norm}\!\left(C(\mathbf{z})\right).
\end{equation}
where $R_{\mathrm{ans}}$ measures answer quality, $C(\mathbf{z})$ denotes token consumption of selected evidence, $\widetilde{C}(\mathbf{z})$ its normalized value, and $\lambda$ controls the balance between answer quality and inference cost.
Optimizing answer quality alone favors richer evidence, whereas increasing $\lambda$ encourages more selective compositions under tighter computation budgets.
This allows \hivemem to adapt evidence usage across different resource constraints without changing the underlying memory representation.

Together, both dimensions of \hivemem operate complementarily.
\emph{Horizontal Memory Expansion} determines which memory episodes to recall, while \emph{Vertical Evidence Composition} determines what information each retrieved memory contributes.
Their combination enables query-adaptive use of multimodal interaction history in both memory coverage and evidence granularity.

\section{Experiments}

\subsection{Experimental Settings}

\textbf{Benchmarks and Metrics.}
We evaluate \hivemem on three multimodal memory benchmarks: Mem-Gallery~\citep{bei2026mem}, WorldMemArena~\citep{liu2026worldmemarena}, and H2HMem~\citep{zhu2026h2hmem}, covering long-horizon interaction histories with both textual and visual information.
For task performance, we report token-level F1 and an LLM-as-a-judge score (LJ), where GPT-4o-mini is used as the evaluator.
To measure inference efficiency, we additionally report total token consumption (Tok.), summing input and output tokens.
Further details on evaluation protocols and judge prompts are provided in Appendix~\ref{app:eval_details}.

\textbf{Baselines.}
We compare \hivemem against two groups of representative memory methods.
For \emph{(1) text-centric memory}, we include \textbf{LightMem}~\citep{fang2025lightmem}, \textbf{SimpleMem}~\citep{liu2026simplemem}, and \textbf{BudgetMem}~\citep{zhang2026learning}, which cover lightweight memory construction, compression, runtime adaptation, and learned memory management.
For \emph{(2) multimodal memory}, we compare against \textbf{M2A}~\citep{feng2026m2a}, \textbf{MIRIX}~\citep{wang2025mirix}, \textbf{MMA}~\citep{lu2026mma}, and \textbf{MemVerse}~\citep{liu2025memverse}, which explicitly model and retrieve information from multimodal interaction histories.

\textbf{Implementation Details.}
We initialize the evidence selector as a lightweight multi-layer perceptron (MLP) and use Proximal Policy Optimization (PPO) as the default reinforcement learning algorithm.
Experiments use Qwen3-VL-4B-Instruct and GPT-5-mini as base models, with GPT-5-mini used only for transfer evaluation (i.e., the selector is only trained on Qwen3-VL-4B-Instruct).
Interaction histories are segmented into episodes of 256 tokens, and Qwen3-VL-Embedding-2B is used for semantic retrieval.
By default, \hivemem retrieves $K_{\mathrm{ret}}=5$ seed memories, expands $K_{\mathrm{exp}}=2$ additional memories, and sets the maximum graph degree to $K_{\mathrm{edge}}=4$, while all baselines use a consistent Top-7 retrieval setting.
Unless otherwise specified, we set $\lambda=0$ and further analyze the effect of budget constraints in Section~\ref{sec:evidence_adaptation}.
We split each benchmark into training, validation, and test sets with a $6{:}2{:}2$ ratio.
Appendix~\ref{app:details} provides additional implementation and evaluation details.

\begin{table*}[t]
\centering
\scriptsize
\caption{
\textbf{Main results on multimodal memory benchmarks.}
We report F1 and LLM-Judge (LJ) for task performance, together with total token consumption (Tok.).
Avg. denotes the average across three benchmarks.
Best and second-best results among multimodal methods are shown in \textbf{bold} and \underline{underlined}; text-centric baselines are included for reference only, as they operate under a different modality setting without image processing during memory-bank construction.
}
\setlength{\tabcolsep}{3pt}
\label{tab:main_results}
\resizebox{\textwidth}{!}{
\begin{tabular}{lcccccccccccc}
\toprule
& \multicolumn{3}{c}{\textbf{Mem-Gallery}}
& \multicolumn{3}{c}{\textbf{WorldMemArena}}
& \multicolumn{3}{c}{\textbf{H2HMem}}
& \multicolumn{3}{c}{\textbf{Avg.}} \\
\cmidrule(lr){2-4}
\cmidrule(lr){5-7}
\cmidrule(lr){8-10}
\cmidrule(lr){11-13}
\textbf{Method}
& \textbf{F1} & \textbf{LJ} & \textbf{Tok.}
& \textbf{F1} & \textbf{LJ} & \textbf{Tok.}
& \textbf{F1} & \textbf{LJ} & \textbf{Tok.}
& \textbf{F1} & \textbf{LJ} & \textbf{Tok.} \\
\midrule

\multicolumn{13}{l}{\cellcolor{datasetbg}\textbf{Qwen3-VL-4B-Instruct}} \\

\rowcolor{groupbg}
\multicolumn{13}{l}{\textit{Text-centric memory}} \\
LightMem
& 27.70 & 38.55 & 0.547
& 21.24 & 45.00 & 0.394
& 16.33 & 26.40 & 0.357
& 21.75 & 36.65 & 0.433 \\

SimpleMem
& 31.76 & 36.27 & 0.462
& 16.31 & 38.50 & 0.598
& 18.30 & 24.11 & 0.433
& 22.12 & 32.96 & 0.498 \\

BudgetMem
& 34.89 & 42.45 & 0.264
& 22.01 & 56.82 & 0.361
& 20.32 & 29.49 & 0.272
& 25.74 & 42.92 & 0.299 \\

\rowcolor{groupbg}
\multicolumn{13}{l}{\textit{Multimodal memory}} \\
M2A
& \underline{47.95} & \underline{62.55} & 11.707
& 14.12 & 37.73 & 24.536
& \underline{23.20} & 31.39 & 19.233
& \underline{28.42} & 43.89 & 18.492 \\

MIRIX
& 39.16 & 51.55 & 6.267
& 20.72 & 60.68 & 12.334
& 23.09 & 31.11 & 7.266
& 27.66 & 47.78 & 8.622 \\

MMA
& 33.51 & 51.55 & 33.034
& \textbf{28.24} & 59.77 & \underline{6.343}
& 22.82 & \underline{33.74} & 7.879
& 28.19 & \underline{48.35} & 15.752 \\

MemVerse
& 29.00 & 44.64 & \underline{5.445}
& 22.06 & \underline{62.73} & 11.916
& 21.36 & 28.89 & \underline{6.074}
& 24.14 & 45.42 & \underline{7.812} \\

\rowcolor{oursbg}
\textbf{\hivemem}
& \textbf{49.44} & \textbf{71.55} & \textbf{0.491}
& \underline{23.66} & \textbf{68.18} & \textbf{0.991}
& \textbf{23.87} & \textbf{34.17} & \textbf{0.492}
& \textbf{32.32} & \textbf{57.96} & \textbf{0.658} \\

\midrule

\multicolumn{13}{l}{\cellcolor{datasetbg}\textbf{GPT-5-mini}} \\

\rowcolor{groupbg}
\multicolumn{13}{l}{\textit{Text-centric memory}} \\
LightMem
& 43.08 & 51.45 & 0.264
& 22.32 & 57.73 & 0.419
& 24.40 & 29.51 & 0.363
& 29.93 & 46.23 & 0.348 \\

SimpleMem
& 33.65 & 48.09 & 0.303
& 21.66 & 54.27 & 0.493
& 21.32 & 32.29 & 0.461
& 25.54 & 44.89 & 0.419 \\

BudgetMem
& 41.11 & 48.09 & 0.224
& 23.61 & 61.59 & 0.411
& 22.36 & 28.82 & 0.355
& 29.02 & 46.17 & 0.330 \\

\rowcolor{groupbg}
\multicolumn{13}{l}{\textit{Multimodal memory}} \\
M2A
& \underline{49.79} & \underline{65.82} & 9.311
& 23.00 & 65.68 & 7.955
& \underline{29.34} & \underline{42.28} & 12.558
& \underline{34.04} & 57.93 & 9.941 \\

MIRIX
& 39.89 & 55.64 & 4.543
& 22.50 & 77.50 & 12.518
& 26.23 & 37.71 & \underline{4.988}
& 29.54 & 56.95 & \underline{7.350} \\

MMA
& 33.87 & 54.55 & \underline{4.228}
& \underline{24.32} & 67.50 & \underline{3.450}
& 25.30 & 37.82 & 19.433
& 27.83 & 53.29 & 9.037 \\

MemVerse
& 45.66 & 63.91 & 8.855
& 24.03 & \underline{78.64} & 10.818
& 29.16 & 39.58 & 12.782
& 32.95 & \underline{60.71} & 10.818 \\

\rowcolor{oursbg}
\textbf{\hivemem}
& \textbf{53.31} & \textbf{72.45} & \textbf{0.367}
& \textbf{24.80} & \textbf{79.32} & \textbf{1.021}
& \textbf{29.71} & \textbf{42.29} & \textbf{0.856}
& \textbf{35.94} & \textbf{64.69} & \textbf{0.748} \\

\bottomrule
\end{tabular}
}

\vspace{2pt}

\begin{minipage}{\textwidth}
\raggedright
\footnotesize
\textit{Note:} Tok. denotes total input and output token consumption, reported in millions (M).
\end{minipage}
\end{table*}

\subsection{Comparison Experimental Results}

\textbf{\hivemem consistently improves task performance across models and benchmarks.}
As shown in Table~\ref{tab:main_results}, text-centric methods generally trail multimodal baselines, highlighting the value of preserving visual information, whereas existing multimodal methods benefit from richer context but often rely on less selective memory utilization. Against this backdrop, \hivemem achieves the best average F1/LJ scores of 32.32/57.96 and 35.94/64.69 with Qwen3-VL-4B-Instruct and GPT-5-mini, respectively, establishing consistent gains over multimodal baselines. The advantage is not driven by a particular benchmark, as \hivemem obtains the best LJ across all three benchmarks with the former backbone and the best F1/LJ across all three with the latter. This improvement supports the benefit of jointly broadening relevant recall and adapting per-memory evidence rather than relying on fixed retrieval and representation schemes. Notably, the latter results use the selector transferred directly from the former without retraining, further indicating cross-model generalization.

\textbf{\hivemem achieves these gains with substantially lower multimodal token consumption.}
Its average consumption is only 0.658M and 0.748M tokens under the two base models, compared with 7.812M and 7.350M for the most efficient competing multimodal baselines, corresponding to reductions of 91.6\% and 89.8\%. This efficiency gap partly arises because existing multimodal methods, despite organizing richer memories through multiple stores, hierarchical structures, or reliability-aware retrieval, still expose substantial retrieved context during answering. In contrast, \hivemem composes evidence separately for each retrieved memory, retaining fine-grained textual or visual information only when useful. This enables substantially lower token consumption without sacrificing performance. For completeness, we also include text-centric methods, whose token usage reflects a different modality setting that omits image processing during memory-bank construction.exclude image processing during memory-bank construction.

Overall, the results show that \hivemem improves multimodal memory performance while reducing inference context. Its consistent gains across benchmarks and backbones further support the benefits of jointly expanding memory coverage and adapting evidence composition.

\subsection{How Does \hivemem Adapt Evidence Composition?}
\label{sec:evidence_adaptation}

\begin{figure*}[t]
    \centering
    \includegraphics[width=\linewidth]{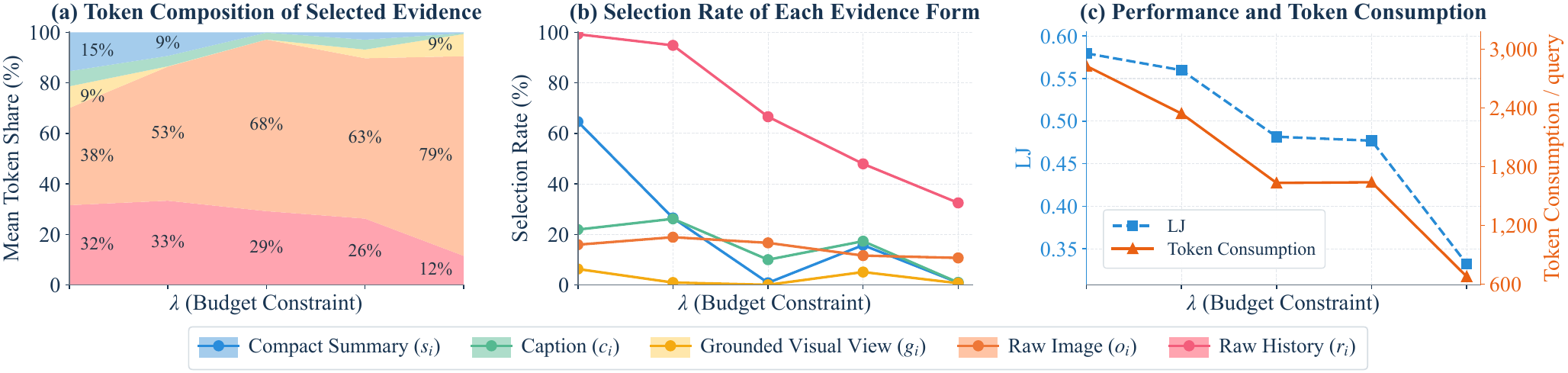}
    \caption{
    \textbf{Evidence composition under varying token-consumption preferences.}
    Results are averaged across Mem-Gallery, WorldMemArena, and H2HMem using Qwen3-VL-4B-Instruct.
    \textbf{(a)} Token shares of selected evidence forms, normalized to 100\%.
    \textbf{(b)} Selection rates of individual evidence forms.
    \textbf{(c)} LJ and token consumption per query under different $\lambda$ values.
}
    \label{fig:evidence_composition}
\end{figure*}

To understand how the learned selector adapts under different budget constraints, we vary $\lambda$ and analyze evidence usage with Qwen3-VL-4B-Instruct, averaging results across the three benchmarks.
Figure~\ref{fig:evidence_composition} shows a clear shift toward more selective compositions as $\lambda$ increases.
The selection rate of raw history drops substantially, while several auxiliary evidence forms are progressively filtered out, indicating that the policy does not reduce cost uniformly across evidence types.
Interestingly, raw images are selected relatively infrequently but contribute an increasingly large fraction of consumed tokens, reflecting their high per-selection cost and the tendency to retain them only when their visual detail is valuable.
This contrast between selection frequency and token share highlights the importance of optimizing evidence utility jointly with consumption.

Figure~\ref{fig:evidence_composition}(c) further shows that varying $\lambda$ produces distinct quality--token-consumption operating points, allowing \hivemem to explicitly control how aggressively evidence is selected under different budget constraints. As $\lambda$ increases, the selector shifts toward more compact evidence compositions and lower token consumption, while smaller values retain richer evidence to prioritize answer quality. Overall, this budget-aware composition enables \hivemem to flexibly adapt to different token budgets without changing the underlying memory system.

\subsection{Ablation Study}

We ablate the key designs of \hivemem using Qwen3-VL-4B-Instruct and report LJ averaged equally across the three benchmarks in Table~\ref{tab:ablation_mean}.
For Horizontal Memory Expansion, \emph{w/o Horizontal Expansion} replaces graph expansion with semantic retrieval while keeping the same number of retrieved memories, \emph{w/o Visual Anchors} removes visual anchors from graph construction, and \emph{w/o IDF Weighting} assigns shared anchors without IDF-based weighting.
For Vertical Evidence Composition, \emph{w/o Vertical Composition} represents every retrieved memory using only its compact summary, while the remaining variants individually remove one evidence form from the composition space.

\textbf{Both horizontal expansion and its relational signals contribute to retrieval quality.}
Disabling Horizontal Memory Expansion reduces LJ from 57.96 to 56.51, showing that retrieving additional memories through cross-episode affinities provides information beyond an equally sized semantic retrieval set.
Removing visual anchors also degrades performance to 57.30, supporting the value of visual cues for connecting multimodal episodes.
Notably, removing IDF weighting causes a larger drop to 56.55, indicating that distinguishing informative anchors from frequently occurring ones is important for constructing useful episode relations.

\textbf{Adaptive evidence composition is particularly important for multimodal memory utilization.}
Replacing Vertical Evidence Composition with compact summaries alone causes the largest degradation, from 57.96 to 44.47, showing that summary-only representations are insufficient for many queries.
Among individual evidence forms, removing raw text and raw images leads to the largest drops, reaching 49.68 and 51.02 LJ, respectively, highlighting the importance of preserving fine-grained information from both modalities.
Compact summaries, captions, and grounded visual views also provide consistent gains when retained.
Together, these results reveal that different evidence forms contribute complementary information across diverse multimodal memory queries, motivating query-adaptive composition rather than relying on compact summaries alone.

\begin{table*}[h]
\centering
\small
\setlength{\tabcolsep}{5pt}
\renewcommand{\arraystretch}{1.1}

\caption{
\textbf{Ablation study of \hivemem} averaged equally
across Mem-Gallery, WorldMemArena, and H2HMEM
using Qwen3-VL-4B-Instruct.
The best LJ score is in \textbf{bold}.
}
\label{tab:ablation_mean}

\resizebox{\linewidth}{!}{%
\begin{tabular}{lclclc}
\toprule

\rowcolor{groupbg}
\multicolumn{2}{c}{\textit{Horizontal Memory Expansion}}
& \multicolumn{4}{c}{\textit{Vertical Evidence Composition}} \\

\cmidrule(lr){1-2}
\cmidrule(lr){3-6}

\textbf{Variant} & \textbf{LJ}
& \textbf{Variant} & \textbf{LJ}
& \textbf{Variant} & \textbf{LJ} \\

\midrule

\rowcolor{oursbg}
\textbf{\hivemem} (Full) & \textbf{57.96}
& \textbf{\hivemem} (Full) & \textbf{57.96}
& \textbf{\hivemem} (Full) & \textbf{57.96} \\

w/o Horizontal Expansion & 56.51
& w/o Vertical Composition & 44.47
& w/o Grounded Visual View & 57.19 \\

w/o Visual Anchors & 57.30
& w/o Compact Summary & 56.86
& w/o Raw Image & 51.02 \\

w/o IDF Weighting & 56.55
& w/o Image Caption & 56.12
& w/o Raw Text & 49.68 \\

\bottomrule
\end{tabular}%
}

\end{table*}

\subsection{Extensive Analysis}

\begin{figure*}[t]
    \centering
    \includegraphics[width=\linewidth]{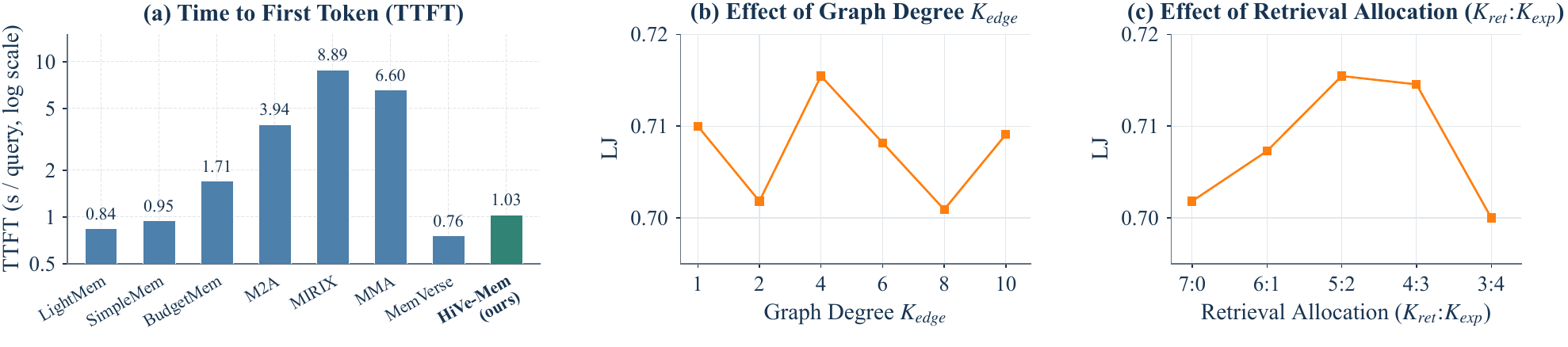}
    \caption{
    \textbf{Latency and hyperparameter sensitivity analysis} with Qwen3-VL-4B-Instruct on Mem-Gallery.
    \textbf{(a)} Time to first token (TTFT) across methods on a logarithmic scale.
    \textbf{(b)} Effect of graph degree $K_{\mathrm{edge}}$ on LJ with default $K_{\mathrm{ret}}:K_{\mathrm{exp}}=5:2$.
    \textbf{(c)} Effect of retrieval allocation on LJ with default $K_{\mathrm{ret}}+K_{\mathrm{exp}}=7$ and $K_{\mathrm{edge}}=4$.
}
    \label{fig:latency_sensitivity}
\end{figure*}

\textbf{Runtime Efficiency Analysis.}
We evaluate time to first token (TTFT) on Mem-Gallery to examine the online overhead introduced by different memory systems.
As shown in Fig.~\ref{fig:latency_sensitivity}(a), several multimodal baselines incur substantially higher TTFT because answering involves additional runtime memory operations, such as agent coordination, memory decomposition, or evidence processing before generation.
In contrast, \hivemem introduces only lightweight graph expansion and evidence composition, resulting in a TTFT of 1.03 seconds.
Although slightly higher than the simplest retrieval-based baselines, this overhead remains marginal and is substantially lower than other multimodal systems in practice.
This suggests that the additional adaptivity of \hivemem can be obtained without materially delaying response generation.

\textbf{Sensitivity Analysis.}
Figures~\ref{fig:latency_sensitivity}(b) and (c) examine the two key hyperparameters of Horizontal Memory Expansion.
Performance remains stable across a wide range of $K_{\mathrm{edge}}$, with the best result at the default $K_{\mathrm{edge}}=4$, indicating that \hivemem is not sensitive to graph sparsity within a reasonable range.
For retrieval allocation, keeping $K_{\mathrm{ret}}+K_{\mathrm{exp}}=7$ reveals a balance between semantic retrieval and graph expansion.
Starting from pure semantic retrieval, increasing $K_{\mathrm{exp}}$ consistently improves LJ and reaches its peak around $5{:}2$, with similar performance at $4{:}3$.
Further increasing the expansion ratio eventually hurts performance, likely because too few semantic seeds provide a narrower and less diverse basis for discovering useful neighbors.
These results show that reserving part of the retrieval budget for graph expansion improves over pure semantic retrieval, while retaining enough semantic seeds remains important for supplying diverse and reliable expansion candidates.
Overall, \hivemem remains robust across a broad range of retrieval configurations.

\textbf{Evidence Selection across Query Types.}
To examine whether \hivemem adapts evidence composition to heterogeneous information needs, we group the native query categories of each benchmark into text-dominant, visual-dominant, and cross-modal queries.
For example, Visual-centric Reasoning (VR) and Visual-centric Search (VS) in Mem-Gallery are treated as visual-dominant, while Test-Time Learning (TTL) represents cross-modal queries.
Similarly, H2HMem distinguishes text-oriented categories such as Temporal Reasoning from visual and cross-modal categories such as Reference \& Evolution Tracking and Cross-modal Related Retrieval.

\begin{figure*}[t]
\centering
\includegraphics[width=\linewidth]{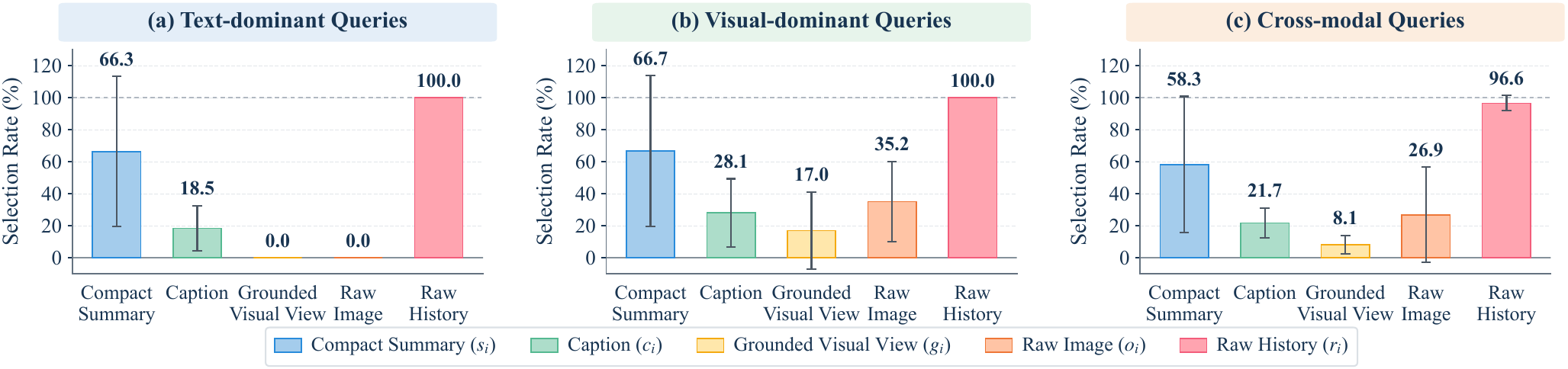}

\caption{
\textbf{Evidence selection across query types.}
Using Qwen3-VL-4B-Instruct at $\lambda=0$, we group the native query categories of each benchmark into
\textbf{(a)} text-dominant,
\textbf{(b)} visual-dominant, and
\textbf{(c)} cross-modal queries, and report the selection rate of each evidence form.
Rates are computed over all retrieved Top-7 memories and averaged equally across Mem-Gallery, WorldMemArena, and H2HMem.
Error bars show $\pm1$ population standard deviation across benchmarks.
}
\label{fig:evidence_selection_by_query_type}
\vspace{-3pt}
\end{figure*}

Figure~\ref{fig:evidence_selection_by_query_type} reveals clear query-dependent evidence patterns.
Text-dominant queries almost never invoke grounded visual views or raw images, whereas visual-dominant and cross-modal queries select visual evidence substantially more often.
At the same time, compact summaries remain frequently selected across all three query types, while raw history provides a stable textual backbone.
Notably, even visual-dominant queries select raw images for only 35.2\% of retrieved memories, indicating that visual information is not indiscriminately exposed whenever a query involves vision. Instead, compressed or grounded evidence can often provide sufficient information, with raw images retained only when additional visual detail is useful.
Cross-modal queries exhibit an intermediate composition that combines textual and visual evidence, further demonstrating that \hivemem adapts evidence at a finer granularity than simply routing queries by modality.
These distinct selection profiles support flexible per-memory evidence composition under heterogeneous query requirements.

\section{Conclusion}

We introduced \hivemem, a multimodal memory framework that improves both memory retrieval and evidence utilization over long text--image interaction histories. \emph{Horizontal Memory Expansion} complements semantic retrieval with cross-episode contextual affinities, while \emph{Vertical Evidence Composition} adaptively selects heterogeneous evidence for each retrieved memory through reinforcement learning. Together, these two dimensions enable more flexible use of multimodal memory across different information needs and inference budgets. Experiments across multimodal memory benchmarks demonstrate consistent improvements over existing methods, with further analyses validating the effectiveness of both memory expansion and adaptive evidence composition.






\subsection*{AI Use Statement}

We used generative AI tools to assist with language polishing and the identification of relevant literature. All AI-assisted text was reviewed and revised by the authors, and identified references were manually verified for correctness and relevance. We take responsibility for the final content of this work.

\subsection*{Reproducibility Statement}

We provide detailed descriptions of the memory construction, Horizontal Memory Expansion, and Vertical Evidence Composition in Secs.~\ref{sec:memory_formation}--\ref{sec:vertical}. Experimental settings, including benchmarks, baselines, model configurations, and evaluation metrics, are described in the main text. Additional details on memory episode construction, training and reward design, and evaluation protocols are provided in Appendices~\ref{app:memory_episode}, \ref{app:training_details}, and \ref{app:eval_details}. We plan to release the source code and implementation upon publication to facilitate reproduction of our results.



\bibliography{iclr2027_conference}
\bibliographystyle{iclr2027_conference}

\clearpage
\appendix
\startcontents[appendix]
\section*{Appendix Contents}
\printcontents[appendix]{}{1}{\setcounter{tocdepth}{2}}
\section*{Appendix}
\section{Implementation Details}
\label{app:details}

\subsection{Memory Episode}
\label{app:memory_episode}

\textbf{Episode Construction.}
We construct memory episodes over approximately fixed-length token spans while respecting natural conversational boundaries. In particular, an episode is not forcibly truncated in the middle of a dialogue or session. If the nominal token boundary falls within an ongoing interaction segment, we extend the episode to the nearest valid boundary. This provides a consistent memory granularity while preserving local conversational continuity. Each raw episode is processed only once by a vision-language model (VLM), which jointly produces its compact summary together with the textual and visual anchors introduced in Sec.~\ref{sec:memory_formation}.

\textbf{Grounded Visual View.}
To obtain a more focused visual representation, we construct the grounded visual view directly from the raw image and its associated interaction context. Specifically, the raw image and the relevant contextual passage are provided to a VLM, which identifies the target object discussed in context and predicts its spatial coordinates. We then crop the corresponding region from the original image to form the grounded visual view. Compared with the raw image, this localized view removes irrelevant background and concentrates visual content around the context-relevant object, yielding a more semantically focused representation. This design also reduces the visual input required by downstream models. \textbf{Depending on the benchmark, grounded visual views reduce token consumption by approximately 20--40\% compared with the raw images.}

Finally, the evidence forms adopted in this work are not intended to define a fixed evidence space. \hivemem treats evidence composition generically and can accommodate additional textual, visual, or derived representations. The raw interaction, compact summary, caption, raw image, and grounded visual view used here serve as representative evidence forms in our default instantiation.

\subsection{Training Details}
\label{app:training_details}

\textbf{Cost Normalization.}
The task reward $R_{\mathrm{ans}}$ is directly computed from answer quality.
For the cost term, we use the token consumption $C(\mathbf{z})$ of the selected evidence as the raw cost.
To reduce sensitivity to its absolute scale, we maintain a sliding window of recent costs and normalize them using robust quantiles.
Letting $Q_5$ and $Q_{95}$ denote the 5th and 95th percentiles of $\sqrt{C}$ within the window, we compute
\begin{equation}
\widehat{C}(\mathbf{z})
=
\operatorname{clip}\!\left(
\frac{\sqrt{C(\mathbf{z})}-Q_5}
{Q_{95}-Q_5},
0,1
\right).
\end{equation}

Since answer-quality and cost signals may have different scales during training, we further align the normalized cost using
\begin{equation}
\alpha
=
\frac{
\operatorname{Std}(R_{\mathrm{ans}})
}{
\operatorname{Std}(\widehat{C})+\epsilon
},
\qquad
\widetilde{C}(\mathbf{z})
=
\alpha\,\widehat{C}(\mathbf{z}),
\end{equation}
where the statistics are computed over recent training samples and $\epsilon$ is a small constant.
The final reward is therefore
\begin{equation}
R
=
R_{\mathrm{ans}}
-
\lambda \widetilde{C}(\mathbf{z}),
\end{equation}
consistent with the formulation in Sec.~\ref{sec:vertical}.

\textbf{PPO Optimization.}
For each query $q$, the per-memory evidence compositions $\mathbf{z}_i$ jointly form the complete evidence composition $\mathbf{z}$.
During PPO training, the selector $f_{\theta}$ parameterizes a factorized policy over the binary evidence decisions through the evidence-specific probabilities predicted in a single forward pass.
Specifically,
\begin{equation}
\pi_{\theta}(\mathbf{z}_i\mid\mathbf{h}_i)
=
\prod_k
\pi_{\theta}(z_{i,k}\mid\mathbf{h}_i),
\end{equation}
where each factor is parameterized by the corresponding selection probability $p_{i,k}$ defined in Sec.~\ref{sec:vertical}.
The composed evidence is then used for downstream answering, yielding the reward $R$ defined above.

Given the behavior policy $\pi_{\theta_{\mathrm{old}}}$ used to collect the rollout, the likelihood ratio of the complete evidence composition is
\begin{equation}
\rho(\theta)
=
\prod_i
\frac{
\pi_{\theta}(\mathbf{z}_i\mid\mathbf{h}_i)
}{
\pi_{\theta_{\mathrm{old}}}(\mathbf{z}_i\mid\mathbf{h}_i)
}.
\end{equation}
We estimate the advantage $\widehat{A}$ from the reward using a learned value baseline and maximize the standard clipped PPO objective
\begin{equation}
\mathcal{J}_{\mathrm{PPO}}
=
\mathbb{E}
\left[
\min\left(
\rho(\theta)\widehat{A},
\operatorname{clip}\!\left(
\rho(\theta),
1-\epsilon_{\mathrm{ppo}},
1+\epsilon_{\mathrm{ppo}}
\right)\widehat{A}
\right)
\right],
\end{equation}
where $\epsilon_{\mathrm{ppo}}$ denotes the PPO clipping coefficient.

\subsection{Evaluation Details}
\label{app:eval_details}

\textbf{Benchmarks and Splits.}
We evaluate on three multimodal memory benchmarks, \textbf{Mem-Gallery}~\citep{bei2026mem}, \textbf{WorldMemArena}~\citep{liu2026worldmemarena}, and \textbf{H2HMem}~\citep{zhu2026h2hmem}, which collectively cover long-horizon interaction histories involving both textual and visual information.
Mem-Gallery evaluates memory over multimodal conversational histories, while WorldMemArena considers memory use over extended agent interactions.
For WorldMemArena, we use only the \emph{Lifelong} subset.
H2HMem further evaluates memory over long multimodal dialogue histories.
Across all benchmarks, we exclude adversarial settings and construct training, validation, and test sets using a $6{:}2{:}2$ split.

\textbf{Model Transfer Evaluation.}
The evidence selector is trained only with Qwen3-VL-4B-Instruct as the base model.
For experiments with GPT-5-mini, we directly transfer the selector weights trained under Qwen3-VL-4B-Instruct and perform evaluation without additional training or adaptation.
This setting allows us to examine whether the learned evidence-selection policy transfers across base models rather than relying on model-specific retraining.

For evaluation, we report token-level F1 and LLM-Judge (LJ), together with input and output token consumption.
GPT-4o-mini is used as the judge model for LJ evaluation across all methods.

\subsection{Hyperparameters}

We provide key hyperparameters used in our work in Table~\ref{tab:key_hyperparameters}.

\begin{table}[t]
\centering
\scriptsize
\caption{Key hyperparameters used in our experiments.}
\label{tab:key_hyperparameters}
\setlength{\tabcolsep}{4pt}
\renewcommand{\arraystretch}{1.08}

\begin{tabular}{
    >{\raggedright\arraybackslash}p{2.35cm}
    >{\centering\arraybackslash}p{1.85cm}
    >{\raggedright\arraybackslash}p{5.65cm}}
\toprule
\textbf{Hyperparameter} & \textbf{Value} & \textbf{Description} \\
\midrule

\rowcolor{groupbg}
\multicolumn{3}{l}{\textit{Embedding and Retrieval}} \\

Embedding model
& Qwen3-VL-Embedding-2B
& Multimodal encoder for queries and memory summaries. \\

Embedding dimension
& 2048
& Dimension of embedding representations. \\

$K_{\mathrm{ret}}$
& 5
& Number of memories obtained by semantic retrieval. \\

$K_{\mathrm{exp}}$
& 2
& Number of memories added by horizontal expansion. \\

\midrule
\rowcolor{groupbg}
\multicolumn{3}{l}{\textit{Evidence Selector}} \\

Architecture
& $2048 \rightarrow 256\times3$
& Three-layer MLP with hidden dimension 256. \\

Selection threshold
& 0.5
& Binary evidence-selection threshold at inference. \\

Initial action probability
& 0.5
& Initial probability of selecting each evidence form. \\

\midrule
\rowcolor{groupbg}
\multicolumn{3}{l}{\textit{PPO Training}} \\

Learning rate
& $3\times10^{-4}$
& Adam learning rate. \\

Clip ratio
& 0.2
& PPO clipping coefficient. \\

Rollout / minibatch
& $16/16$
& Rollout and optimization minibatch sizes. \\

Update epochs
& 4
& PPO updates per rollout batch. \\

Training epochs
& 6
& Total training epochs. \\

$\gamma$ / GAE $\lambda$
& $1.0/1.0$
& Discount factor and GAE coefficient. \\

Value / entropy coef.
& $0.5/0.01$
& Value-loss and entropy coefficients. \\

Max gradient norm
& 1.0
& Gradient clipping threshold. \\

\midrule
\rowcolor{groupbg}
\multicolumn{3}{l}{\textit{Reward}} \\

Quality reward
& F1
& Answer-quality reward. \\

\midrule
\rowcolor{groupbg}
\multicolumn{3}{l}{\textit{Inference}} \\

Temperature
& 0.0
& Deterministic generation. \\

Max output tokens
& 512
& Maximum generation length per invocation. \\

\bottomrule
\end{tabular}
\end{table}

\section{Extensive Experiments}
\label{app:exp}

We further examine the behavior of \hivemem under the GPT-5-mini transfer setting.
Unless otherwise specified, the evidence selector trained with Qwen3-VL-4B-Instruct is directly transferred without retraining.
These experiments complement the analyses in the main text by examining whether adaptive evidence composition remains effective across base models.

\subsection{Evidence Composition under Transfer Evaluation}

\begin{figure*}[h]
    \centering
    \includegraphics[width=\linewidth]{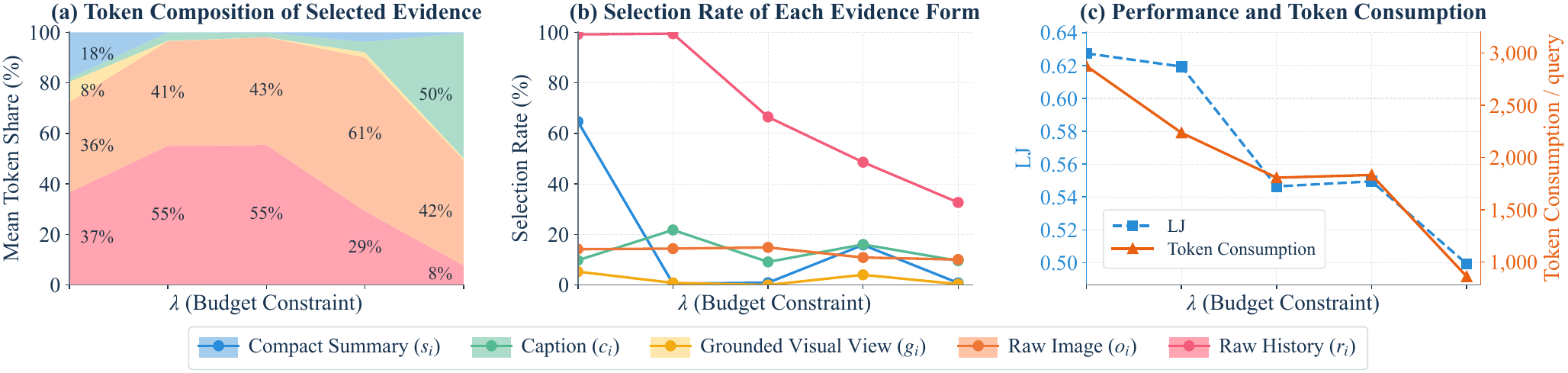}
    \caption{
        \textbf{Evidence composition under varying token-consumption preferences with GPT-5-mini.}
        Results are averaged across Mem-Gallery, WorldMemArena, and H2HMem using the transferred evidence selector.
        \textbf{(a)} Token shares of selected evidence forms, normalized to 100\%.
        \textbf{(b)} Selection rates of individual evidence forms.
        \textbf{(c)} LJ and token consumption per query under increasing $\lambda$.
        Displayed settings are equally spaced in increasing $\lambda$ order.
    }
    \label{fig:evidence_composition_gpt5mini}
\end{figure*}

Figure~\ref{fig:evidence_composition_gpt5mini} shows that the adaptive behavior observed with Qwen3-VL-4B-Instruct persists after transfer to GPT-5-mini.
As $\lambda$ increases, the selector does not uniformly reduce all evidence forms, but progressively changes the composition itself.
Raw history is selected much less frequently under stronger constraints, while visual evidence remains selectively retained when useful.
Although raw images are chosen for only a small fraction of retrieved memories, they account for a substantial share of consumed tokens whenever selected, again reflecting the heterogeneous cost of different evidence forms.
Meanwhile, Fig.~\ref{fig:evidence_composition_gpt5mini}(c) exhibits distinct LJ--token-consumption operating points as $\lambda$ varies.
This demonstrates that the transferred selector preserves controllable evidence allocation and can adjust inference behavior to different token budgets without retraining.

\subsection{Evidence Selection across Query Types}

\begin{figure*}[h]
    \centering
    \includegraphics[width=\linewidth]{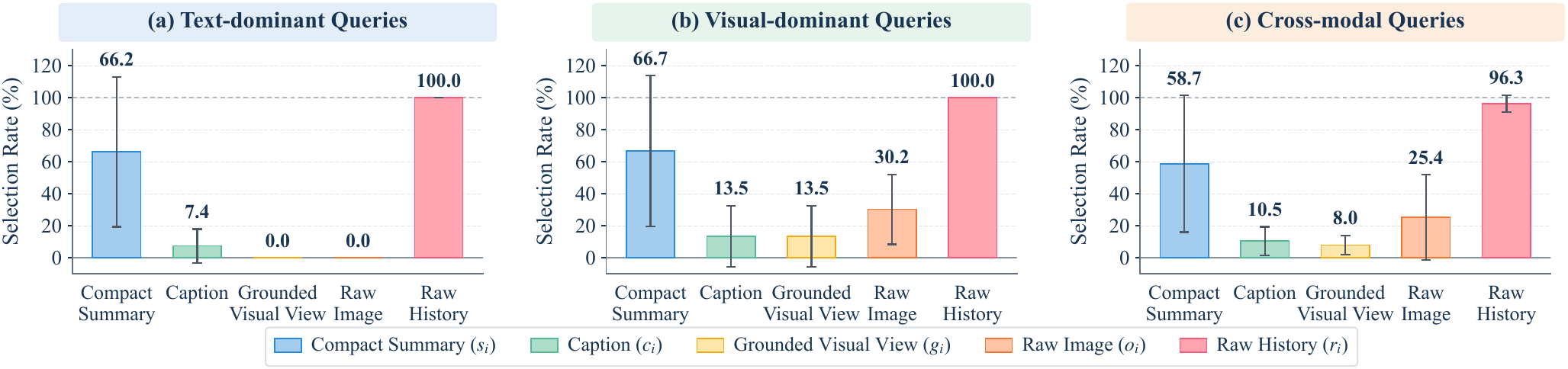}
    \caption{
        \textbf{Evidence selection across query types with GPT-5-mini.}
        Native query categories are grouped into
        \textbf{(a)} text-dominant,
        \textbf{(b)} visual-dominant, and
        \textbf{(c)} cross-modal queries.
        We report evidence-selection rates using the transferred selector at $\lambda=0$.
        Rates are averaged equally across Mem-Gallery, WorldMemArena, and H2HMem, with error bars showing $\pm1$ population standard deviation across benchmarks.
    }
    \label{fig:evidence_selection_query_gpt5mini}
\end{figure*}

Figure~\ref{fig:evidence_selection_query_gpt5mini} further reveals clear query-dependent evidence preferences.
For text-dominant queries, grounded visual views and raw images are almost never selected, indicating that \hivemem avoids unnecessary visual processing when textual evidence is sufficient.
In contrast, visual-dominant and cross-modal queries invoke both grounded visual views and raw images substantially more often, with raw-image selection reaching 30.2\% and 25.4\%, respectively.
Compact summaries remain frequently selected across all query types, while raw history serves as a consistently useful textual source.
Importantly, visual evidence is still selected far below 100\% even for visual-dominant queries, suggesting that relevant visual information is often already captured by compact or derived representations and that raw images are retained only when additional visual detail is needed.
The consistent modality-dependent behavior under GPT-5-mini further supports the transferability and flexibility of per-memory evidence composition.

\section{Limitations and Future Work}
\label{app:limitations-future-work}

Although \hivemem demonstrates consistent effectiveness across multimodal memory benchmarks, several extensions remain open.
First, our experiments instantiate evidence composition with a representative set of textual and visual forms. Future work could incorporate richer derived evidence, such as structured visual descriptions or task-specific representations, within the same framework.
Second, we adopt a lightweight MLP as the evidence selector for efficiency. Exploring alternative selector architectures may further improve evidence composition under different model and task settings.
Finally, our evaluation considers two representative base models. Extending the study to a broader range of model families and scales would provide a more comprehensive understanding of the transferability of \hivemem.

\section{Broader Impact}
\label{app:broader-impact}

As multimodal agents evolve from short-lived question answering systems toward persistent interactive systems, the ability to selectively reuse long-term interaction history may become increasingly important. \hivemem provides a step toward memory systems that do not treat stored multimodal information as a fixed representation, but instead adapt both what should be recalled and how retrieved information should be exposed for downstream reasoning. This perspective could support a broad range of long-horizon applications, including personalized assistants, multimodal copilots, embodied agents, and interactive systems that continuously accumulate textual and visual context over time.

More broadly, adaptive memory utilization may become a fundamental capability for scalable agent systems. As interaction histories grow, processing all stored information uniformly becomes increasingly impractical. Mechanisms that dynamically recover relevant memories and compose only the necessary evidence could enable agents to maintain richer histories while controlling inference overhead. We hope \hivemem encourages future research toward more flexible multimodal memory architectures that integrate additional modalities, richer memory representations, and increasingly adaptive mechanisms for organizing and reusing long-term interaction history.

\section{Prompts}
\label{app:prompt-display-templates}

\subsection{Memory Construction Prompt}

\begin{tcolorbox}[promptbox, title={Task Prompt Template}]
\textbf{Role.}
You convert one conversation chunk and its attached images into exactly one
structured memory record.

\textbf{Input.}
\texttt{\detokenize{<conversation chunk>}},
optionally \texttt{\detokenize{<attached original images>}}, and
optionally \texttt{\detokenize{<user name>}}. The user name may only be used
to resolve references to the user; no other profile facts may be used as
summary or \texttt{Ti} evidence.

\textbf{Instruction.}
Produce exactly one complete summary for the entire chunk, together with:

\begin{itemize}
    \item \texttt{Ti}: attributes explicitly stated in the chunk text;
    \item \texttt{Vi}: attributes directly observable in the attached images.
\end{itemize}

The summary may combine relevant information from the text and images.
Preserve names, numbers, dates, decisions, reasons, and image details needed
for future question answering. If the chunk has a date, begin the summary
with \texttt{\detokenize{On <session date>, }}. Resolve relative dates when
the chunk provides sufficient information.

Treat \texttt{Ti} and \texttt{Vi} as sparse retrieval properties. Include
only concrete and distinctive attributes useful for linking the current
chunk to other memories.

\textcolor{pppromptamber}{\textbf{Constraint.}}
Write exactly one non-empty summary and never split the chunk into multiple
memory records. Do not add unsupported information.

A \texttt{Ti} attribute must be supported by words in the current chunk.
Do not derive \texttt{Ti} from images or user-profile information. Do not
convert questions, uncertain suggestions, conversation mechanics, or
speaker metadata into asserted attributes.

A \texttt{Vi} attribute must be directly observable in an attached image.
Do not derive \texttt{Vi} from the chunk text, image captions, or profile.
Do not visually infer occupation, ownership, preference, causality, identity,
date, or other non-observable facts. If no image is attached, return
\texttt{\detokenize{"Vi":[]}}.

Each attribute value must be an array of strings containing at most eight
distinct values. Each of \texttt{Ti} and \texttt{Vi} may contain at most
16 attribute objects. Merge values that share the same entity and attribute,
and do not repeat the same fact under multiple keys.

Allowed \texttt{Ti} attributes are:
\texttt{relation}, \texttt{preference}, \texttt{occupation},
\texttt{trait}, \texttt{age}, \texttt{species}, \texttt{breed},
\texttt{owner}, \texttt{appearance}, \texttt{skill}, \texttt{status},
\texttt{category}, \texttt{color}, \texttt{style}, \texttt{material},
\texttt{use}, \texttt{kind}, \texttt{location}, \texttt{feature},
\texttt{role}, \texttt{date}, \texttt{participants}, \texttt{shape},
\texttt{texture}, \texttt{pattern}, \texttt{count},
\texttt{visible\_state}, \texttt{action}, \texttt{pose},
\texttt{position}, \texttt{spatial\_relation}, and \texttt{ocr\_text}.

Allowed \texttt{Vi} attributes are:
\texttt{color}, \texttt{appearance}, \texttt{shape}, \texttt{material},
\texttt{texture}, \texttt{pattern}, \texttt{count},
\texttt{visible\_state}, \texttt{action}, \texttt{pose},
\texttt{position}, \texttt{spatial\_relation}, and \texttt{ocr\_text}.

If the same attribute is independently supported by both sources, include it
once in \texttt{Ti} and once in \texttt{Vi}; never copy evidence from one
source to the other.

\textbf{Output.}
Return exactly one valid JSON object:

\texttt{\detokenize{
{"summary":"<one complete summary of the entire chunk>",
"Ti":[{"entity":"<optional entity name>",
"attribute":"<allowed textual attribute>",
"value":["<one or more values>"]}],
"Vi":[{"entity":"<optional visible entity>",
"attribute":"<allowed visual attribute>",
"value":["<one or more values>"]}]}
}}

Do not output a JSON array, multiple objects, multiple summaries,
\texttt{MEMORY\_ITEM} or \texttt{ENTITIES} lines, Markdown, or explanatory
text.
\end{tcolorbox}

\subsection{QA Prompts}
\label{app:qa-prompt}

\begin{tcolorbox}[promptboxgreen,
    title={QA Instructions: Mem-Gallery}]

Answer from the provided multimodal conversation memory.
Ground the answer in that memory, prefer the latest information
when details change over time, and keep the response concise
but complete. When the question asks for images, preserve the
exact \texttt{image\_id} values.

\medskip
\textit{Additional user-message constraint for conflict detection:}

The text inside
\texttt{\detokenize{<answer>...</answer>}}
must be exactly ``Yes.'' or ``No.''

\medskip
\textit{Additional user-message constraint for visual search:}

Return the exact matching \texttt{image\_id} value or values.
If several apply, sort them in ascending order and separate
them with commas.

\end{tcolorbox}

\begin{tcolorbox}[promptboxgreen,
    title={QA Instructions: WorldMemArena}]

Answer from the provided retrieved memories.
Synthesize multiple entries when needed, use timestamps and the
most recent evidence to resolve conflicts, and ground every
claim in the memories. Convert relative dates using the memory
timestamps. When images are requested, preserve exact
\texttt{image\_id} values.
Keep the answer concise; if no relevant information is present,
the answer text must be exactly ``Not mentioned in memory.''

\end{tcolorbox}

\begin{tcolorbox}[promptboxgreen,
    title={QA Instructions: H2HMem}]

You are a memory testing system.
\texttt{\detokenize{<question-type instruction>}}
Answer directly in English without reasoning, using no more
than 100 words.

\medskip
\textit{Additional system instruction for conflict detection:}

The answer text must be strictly either Yes or No.

\medskip
\textit{Additional system instruction for answer refusal:}

If the information is absent, the answer text must be exactly
``Not mentioned.''

\end{tcolorbox}

For H2HMem, the question-type instruction is selected from
the following:

\begin{tcolorbox}[promptboxgreen,
    title={H2HMem Question-Type Instructions}]

\textbf{Unimodal precise recall.}
Accurately recall the requested information from the
conversation memory.

\textbf{Cross-modal related retrieval.}
Use related textual and visual information from the
conversation memory.

\textbf{Knowledge resolution.}
Resolve knowledge consistently across the conversation memory.

\textbf{Temporal reasoning.}
Reason about temporal relationships in the conversation memory.

\textbf{Multimodal causal reasoning.}
Use textual and visual evidence for the requested causal reasoning.

\textbf{Reference \& evolution tracking.}
Track references and how they evolve across the conversation memory.

\textbf{Test-time learning.}
Use knowledge introduced in the conversation memory to answer
the question.

\textbf{Conflict detection.}
Determine whether the information in the question conflicts
with the conversation memory.

\textbf{Answer refusal.}
Determine whether the question can be answered from the
conversation memory.

\end{tcolorbox}

\subsection{LLM Judging Prompts}
\label{app:judge-prompt}

\paragraph{Mem-Gallery and H2HMem.}

These benchmarks share the following five-level rubric.
In the output-format example, the reasoning placeholder is empty
for Mem-Gallery and \texttt{\detokenize{<short explanation>}}
for H2HMem.

\begin{tcolorbox}[
    promptboxred,
    breakable,
    title={Judge Prompt: Mem-Gallery and H2HMem}
]

You are an impartial judge evaluating the memory capabilities
of an AI assistant with the question-answering task.

Your task is to compare the Assistant's Answer against the
Ground Truth and assign a score of 0, 0.25, 0.5, 0.75, or 1.

\textbf{Scoring Rubric.}

\textbf{Score 0 (Incorrect / Miss):}
\begin{itemize}
    \item The answer contradicts the Ground Truth.
    \item For Yes/No questions: The answer has the wrong polarity
    (e.g., says ``Yes'' when Ground Truth is ``No'').
    \item For Open-ended questions: The answer provides factually
    wrong information or hallucinations.
    \item The assistant fails to provide the required information.
\end{itemize}

\textbf{Score 0.25 (Poor / Tangential):}
\begin{itemize}
    \item The answer touches on the topic but misses the
    \textbf{core entity} or key value required.
    \item The answer contains a mix of minor correct details and
    \textbf{significant hallucinations} or wrong associations.
    \item The answer is excessively vague to the point of being
    useless (e.g., answering ``a dog'' instead of
    ``a golden retriever'').
\end{itemize}

\textbf{Score 0.5 (Partial / Vague):}
\begin{itemize}
    \item The answer is technically correct, but lacks confidence
    or is incomplete.
    \item The answer captures the \textbf{main entity or concept}
    correctly but misses a part of the required supporting details.
    \item For Yes/No questions: The polarity is correct, but the
    reasoning is flawed (if have), or the assistant is uncertain
    (e.g., ``I think it might be Yes'').
    \item For Open-ended questions: The answer is too general or
    misses key adjectives/details present in the Ground Truth.
\end{itemize}

\textbf{Score 0.75 (Good / Minor Imperfection):}
\begin{itemize}
    \item The answer is largely accurate and captures the core
    information confidently.
    \item It misses only \textbf{minor details}
    (e.g., specific adjectives or secondary details)
    that do not alter the main truth.
    \item The answer contains all the correct information but
    includes unnecessary ``fluff'' or slight conversational filler
    that reduces precision.
\end{itemize}

\textbf{Score 1 (Correct / Exact):}
\begin{itemize}
    \item The answer is accurate, precise, and confident.
    \item For Yes/No questions: The polarity matches the
    Ground Truth perfectly.
    \item For Open-ended questions: The answer contains
    \textbf{all} the core information and necessary details
    required by the Ground Truth without hallucinations.
\end{itemize}

\textbf{Input Data.}

Ground Truth:
\texttt{\detokenize{<reference answer(s)>}}

Assistant Answer:
\texttt{\detokenize{<predicted answer>}}

\textbf{Output Format.}

Output strictly in the following JSON format:
\begin{flushleft}
\small\ttfamily
\{"score": <0, 0.25, 0.5, 0.75, or 1>,\par
\hspace*{1em}"reasoning": "<reasoning placeholder>"\}
\end{flushleft}

\end{tcolorbox}

\paragraph{WorldMemArena.}

The evaluator assigns a score of 1 to \texttt{Correct}
and 0 to \texttt{Hallucination} or \texttt{Omission}.
This evaluates answer correctness, not evidence coverage.

\begin{tcolorbox}[
    promptboxred,
    breakable,
    title={Judge Prompt: WorldMemArena}
]

You are an \textbf{evaluation expert for AI memory system
question answering}.

Based \textbf{only} on the provided \textbf{Reference Answer},
strictly evaluate the \textbf{accuracy} of the
\textbf{Memory System Response}.
Classify it as one of \textbf{Correct},
\textbf{Hallucination}, or \textbf{Omission}.
Do \textbf{not} use any external knowledge or subjective inference.

\textbf{Evaluation Criteria.}

\textbf{1. Correct}
\begin{itemize}
    \item The response accurately answers the question and is
    \textbf{semantically equivalent} to the Reference Answer.
    \item No contradictions with the Reference Answer.
    \item Synonyms, paraphrasing, and reasonable summarization
    are acceptable.
\end{itemize}

\textbf{2. Hallucination}
\begin{itemize}
    \item The response includes information that
    \textbf{contradicts} the Reference Answer.
    \item When the Reference Answer is \emph{unknown/uncertain},
    yet the response provides a specific fact.
\end{itemize}

\textbf{3. Omission}
\begin{itemize}
    \item The response is \textbf{incomplete} compared to the
    Reference Answer.
    \item It states ``don't know'' or ``no related memory''
    even though the Reference Answer supplies the answer.
    \item For multi-element answers, missing \textbf{any}
    element counts as Omission.
\end{itemize}

\textbf{Priority Rules.}
\begin{itemize}
    \item Both missing info AND fabricated info
    \(\rightarrow\) \textbf{Hallucination}.
    \item No fabrication but missing info
    \(\rightarrow\) \textbf{Omission}.
    \item Fully equivalent
    \(\rightarrow\) \textbf{Correct}.
\end{itemize}

\textbf{Information.}

Reference Answer:
\texttt{\detokenize{<reference answer(s)>}}

Memory System Response:
\texttt{\detokenize{<predicted answer>}}

\textbf{Output.}
\begin{flushleft}
\small\ttfamily
\{\par
\hspace*{1em}"reasoning": "Concise evaluation rationale",\par
\hspace*{1em}"evaluation\_result":\par
\hspace*{2em}"Correct | Hallucination | Omission"\par
\}
\end{flushleft}

\end{tcolorbox}

\end{document}